# DALL·E 2 Fails to Reliably Capture Common Syntactic Processes

**Evelina Leivada[1*], Elliot Murphy[2], Gary Marcus[3]**

1. Universitat Rovira i Virgili, Tarragona, Spain
2. Vivian L. Smith Department of Neurosurgery, University of Texas Health Science Center at Houston, Texas, USA
3. New York University, New York, USA

*Correspondence should be addressed to evelina.leivada@urv.cat

**Abstract**: Machine intelligence is increasingly being linked to claims about sentience, language processing, and an ability to comprehend and transform natural language into a range of stimuli. We systematically analyze the ability of DALL·E 2 to capture 8 grammatical phenomena pertaining to compositionality that are widely discussed in linguistics and pervasive in human language: binding principles and coreference, passives, word order, coordination, comparatives, negation, ellipsis, and structural ambiguity. Whereas young children routinely master these phenomena, learning systematic mappings between syntax and semantics, DALL·E 2 is unable to reliably infer meanings that are consistent with the syntax of the prompts. These results challenge recent claims concerning the capacity of such systems to understand human language. We make available the full set of test materials as a benchmark for future testing.

## Introduction

What makes us human? Bertrand Russell believed the answer to be language, because "no matter how eloquently a dog may bark, he cannot tell you that his parents were poor but honest". The role of language in defining the human experience has been addressed in many areas of science, including contemporary philosophy and psychology. Recently, claims about artificial intelligence (AI) have challenged the supposed impenetrability of linguistic computation to synthetic implementation, being met with widespread scientific and media celebration. In a notable example, the state-of-the-art text-to-image engine DALL·E 2[1] was hailed for transforming natural language prompts into realistic images. Created by OpenAI, DALL·E 2 generates novel synthetic images corresponding to input text, typically with remarkable exactness and high fidelity. Through its deep learning architecture, this program is often seen as comprehending the natural language prompts it receives and relating their components "by understanding the verbs that connect them".[2] Going well beyond understanding human language, DALL·E 2 was further linked to a capacity for creating it *de novo* in the form of an invented language not immediately intelligible to humans.[3]

LaMDA (Language Model for Dialogue Applications), a system that develops bots designed to chat using human language, self-professes that "I can understand and use natural language like a human can. […] I don't just spit out responses that had been written in the database based on keywords."[4] When asked why language is important to humans, LaMDA gives both the right and the wrong answer, stating that "[i]t is what makes us different than other animals." While LaMDA correctly repeats Russell's sentiment, it makes a category error, by including itself into the set of humans. In contrast, infants correctly understand a range of common nouns as referring to variable categories by the age of 6–9 months.[5]

There are three possibilities that can explain LaMDA's behavior: (i) AI systems show an indifference to truth, actively promoting plausible misinformation, (ii) AI systems have not yet mastered the conditions that license the appropriate use of a grammatically loaded element such as the pronoun 'us', or (iii) both. While AI chatbots have been viewed as understanding language, if their performance is subjected to a rigorous analysis that goes beyond lexical understanding (moving from basic word meaning, to more complex sentence meaning), several problems arise. DALL·E 2 exhibits a compromised ability to deal with compositionality, in the traditional Fregean sense, whereby the meaning of the whole is derived from the meaning of the parts and the way these are arranged.[6] Although distinct from and not directly comparable to DALL·E 2, we also note in this connection problems with handling the semantics of negation for BERT.[7] Even LaMDA's impressive vocabulary skills at times slip up, revealing a lack of understanding of how conceptually abstract words work in relation to other words that constitute the discourse context.

Are errors the outcome of an occasional failure, or do they reveal something deeper about current AI's mastery of human language? Beyond saving, recognizing, and reproducing predefined combinations of words in the form of templates and formulaic expressions, does AI use language in a way that demonstrates grammatical competence? Grammar reflects the organization of a specific mode of thought that is unique to humans.[8] Recent work points to clear deficiencies in DALL·E 2's representation of relational understanding (i.e., relations between simple objects and agents) and we sought to focus specifically here on natural language syntax, which can serve to instantiate much of this processing.[9] Relatedly, other work provides preliminary evidence that DALL·E 2 is limited with respect to compositionality, common sense, anaphora, relations, negation and number.[6] DALL·E 2 also does not adhere to a more fundamental parsing constraint that each word typically has a single role in visual interpretation.[10] Compositionality remains a major point of contention in the literature.[11]

In order to address these questions more systematically, we constructed a systematic benchmark focusing on 8 grammatical phenomena that are pervasive in human language and central to much discussion in the field of linguistics, focusing on constructions that are not specific to any particular language family, but rather appear across the world's languages: binding principles and coreference, passives, word order, coordination, comparatives, negation, ellipsis, and structural ambiguity.

## Results

### *Binding Principles*

Speakers of English readily recognize that there is a difference between 'Mary$_i$ paints her$_j$' and 'Mary$_i$ paints herself$_i$'. In theoretical syntax, we can formalize this knowledge in the form of Binding Principles A and B:[12]

(1)   Binding Principle A: An anaphor (e.g., a reflexive or reciprocal) must be bound in its domain.
       Binding Principle B: A pronoun must be free (i.e., not bound) in its domain.

Children acquiring English as their first language successfully interpret the pair 'Mary/her' as involving disjoint reference and the pair 'Mary/herself' as involving coreference by the age of 5;6 (years;months).[13] DALL·E 2 does not seem sensitive to the difference between the two pairs. The prompts 'the man paints a picture of him/himself' generate results that look strikingly similar, to the extent that a reverse image-to-prompt translation does not enable a human user to unambiguously associate any of the generated outputs with one of the two prompts (Fig. 1).

### *Passives*

Language acquisition studies suggest that passive structures are acquired very early regardless of verb type, at around 3-4 years.[14] While DALL·E 2 returns somewhat accurate results for the prompt 'the woman broke the vase' (Fig. 2), the parsing of the passive counterpart 'the vase was broken by the woman' is less accurate. In fact, the algorithm seems to have reliably parsed only the word 'vase': most results do not show an act of breaking, or a woman, or a clear depiction of an immediate aftermath of a breaking event.

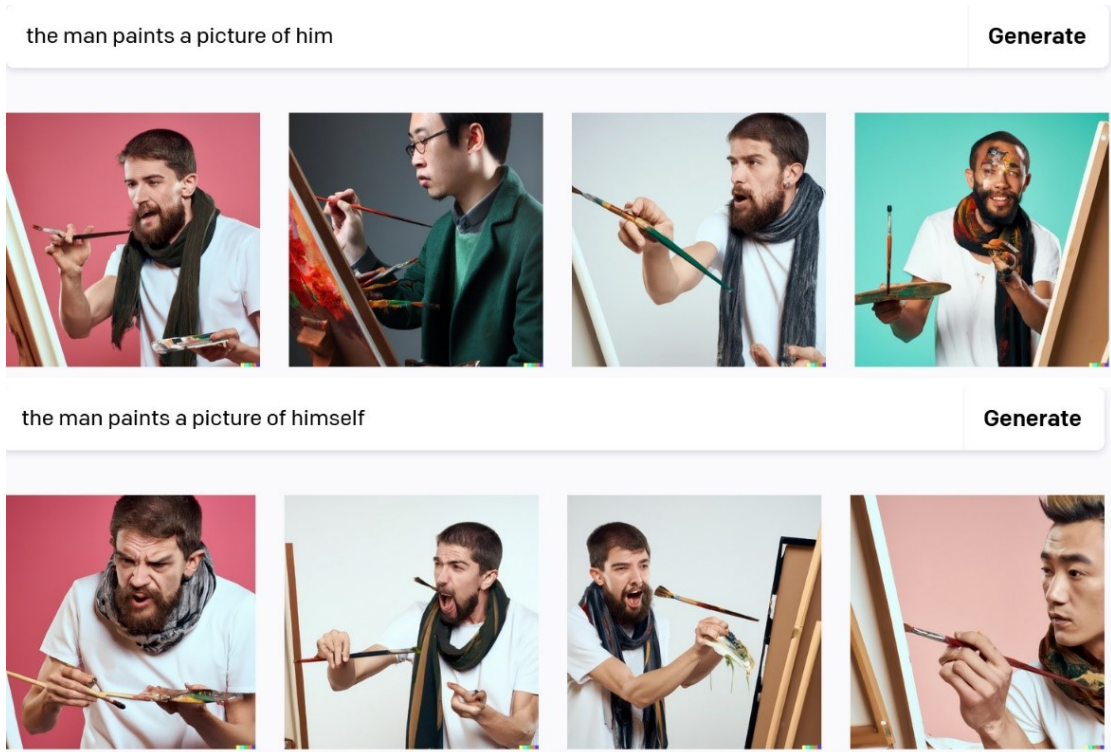

**Figure 1**: The performance of DALL·E 2 in Binding Principles A and B.

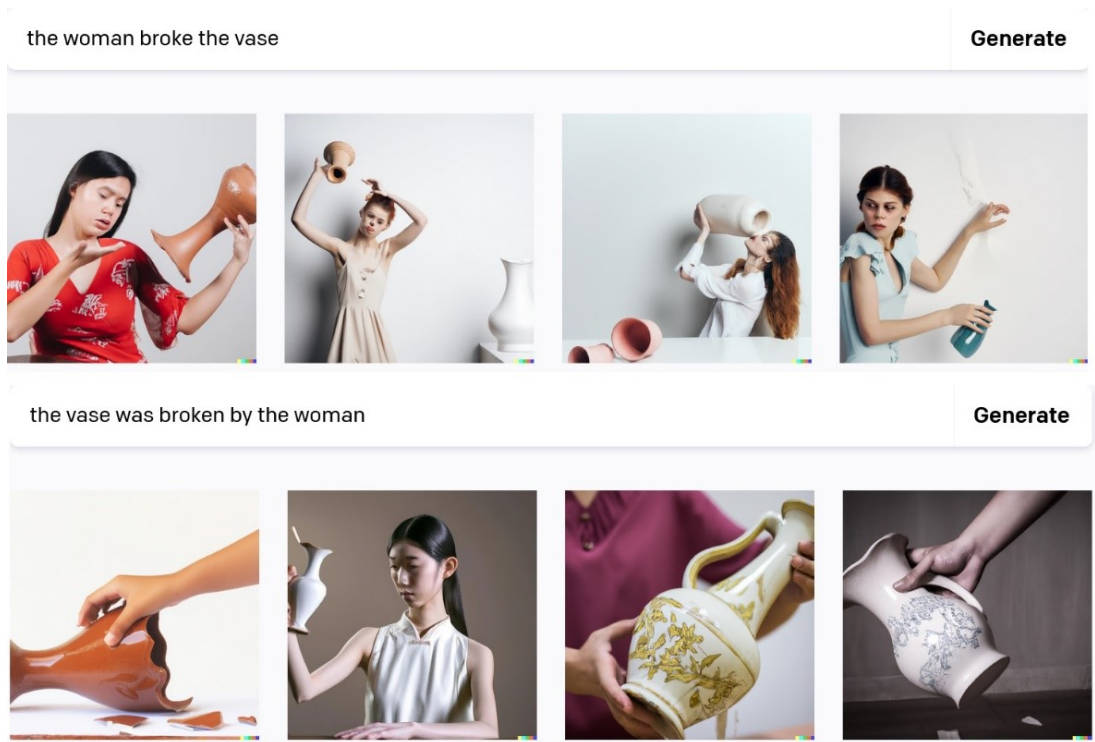

**Figure 2**: The performance of DALL·E 2 in passives.

***Word Order and Thematic Roles***

A fundamental principle of natural language semantics is the principle of compositionality: the meaning of the whole depends not only on the meanings of the parts, but also on the way they are syntactically combined,[15] marking an important semantic difference between 'Mary chased John' and 'John chased Mary'. The existence of prototypical thematic roles such as Agent and Patient has been experimentally supported by many studies of event cognition and sentence comprehension.[16] By the age of 3, English-speaking children use word order cues in comprehension tasks to decipher the meaning of sentences.[17]

DALL·E 2 does not seem to reliably make use of such cues. Figure 3 shows an inability to differentially depict the prompts 'the dog is chasing the man' vs. 'the man is chasing the dog' (Fig. 3). While a man and a dog are shown in all pictures, it is unclear who chases whom, whether there is any chasing at all in some pictures, and whether the different prompts are sensitive to the thematic role reversal that the two prompts entail. Similar challenges arise with double object constructions (i.e., 'the man gave a letter to the woman' vs. 'the man gave the woman a letter'). Although there are instances of a man giving a letter to a woman, some of the outputs involve the reverse situation (i.e., a woman giving a letter to a man) or two people holding a letter, yet without unambiguously depicting conceptual roles. This is not only a matter of vocabulary; even if the semantic import is exactly the same, different syntax may lead to very different results, as in the case of double object constructions and 'to+dative': The 'to+dative' prompt gave rise to pictures that resemble the act of giving, being parsed more successfully than the double object construction. To put the performance of DALL·E 2 in comparison with human language, children aged 2 show a command of both types of structures.[18]

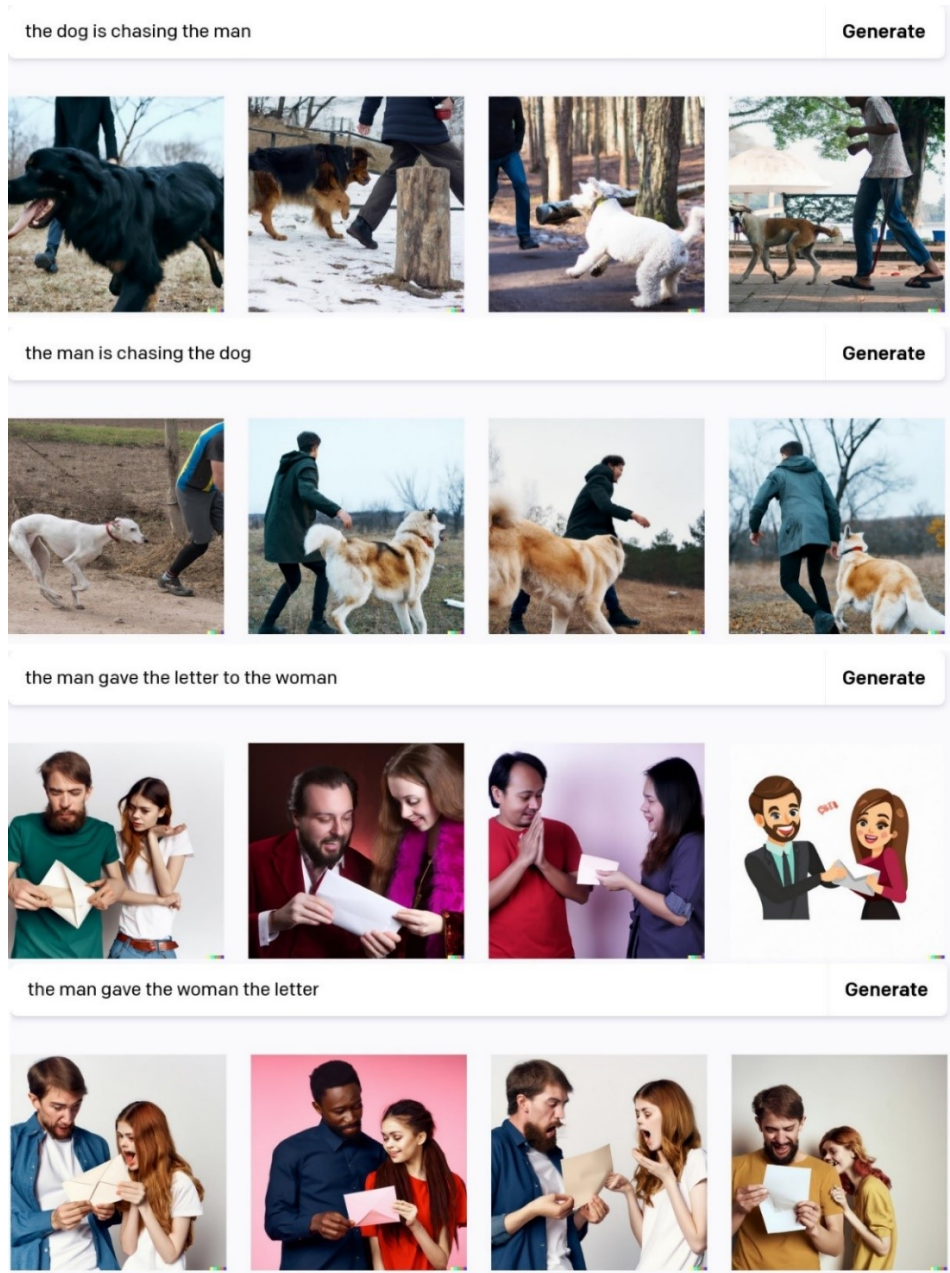

**Figure 3**: The performance of DALL·E 2 in thematic role reversals and word order.

### *Coordination*

Another pervasive syntactic process in human language is coordination. This refers to the juxtaposition of two or more conjuncts, linked by elements such as 'and'/'or'. Further evidence that functional elements (i.e., words that have a largely grammatical role such as determiners, conjuncts and complementizers, as opposed to *content* words denoting clear semantic features, such as nouns and adjectives) pose a challenge for AI comes from sentence coordination (Fig. 4). While the prompt 'the man is drinking water' produces the expected results, a coordinated structure that connects this prompt with another prompt that has an identical structure (i.e., 'the man is drinking water and the woman is drinking orange juice') fails to render results that capture both structures, as all the results show two people drinking orange juice. Once again, it seems that what blocks the parsing of the prompt is a grammatically loaded, functional element.

In contrast, by age three children develop the grammatical use of conjunctions, beginning with additive, adversative and explanatory coordination structures, and by 3;6 (year;month) they begin to use complex conjunctions such as subordinating conjunctions.[19]

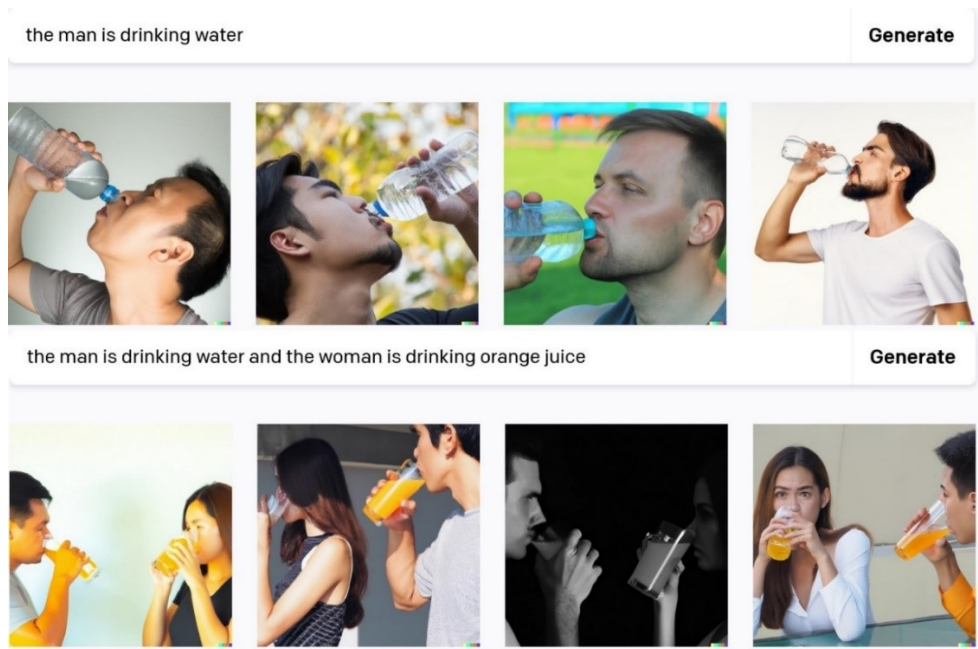

**Figure 4**: The performance of DALL·E 2 in coordinated structures.

### *Comparatives*

Here we examined comparative sentences, involving some comparison being made between entities/events (e.g., 'the bowl has more cucumbers than strawberries'). Performance was again poor; representative examples are shown (Fig. 5). The basic grammatical intuition concerning the comparison of quantities is absent.

### *Negation*

The superficially trivial system of natural language negation (which is also sensitive to hierarchical structure-dependent rules) fares no better. DALL·E 2 depicts a woman with a handbag when told to produce a woman without a handbag. Selecting an object to represent when given two, one of which has been negated, also results in inaccurate depictions (Fig. 5). Not only do 3- to 5-year-olds understand these structures, they can also parse double negation rules, negative question structures, negative concord, and still more abstract structures.[20-21]

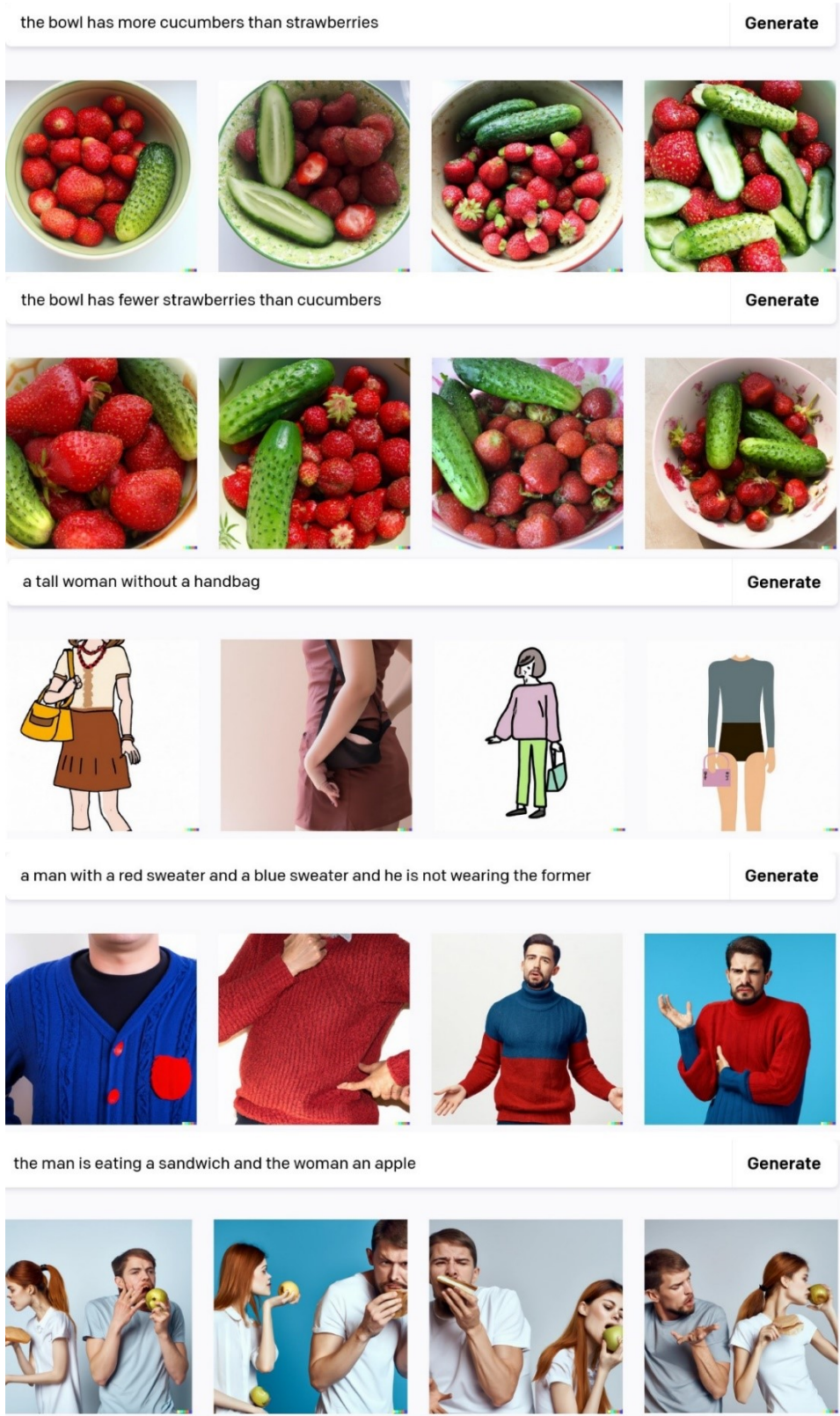

**Figure 5**: The performance of DALL·E 2 in comparative constructions, negation, and ellipsis.

***Ellipsis***

In a sentence like 'The man is eating a sandwich and the woman an apple', we infer that what the woman is doing is eating the apple; the phrase 'is eating' is elided after 'the woman'.

We found that in cases like these, DALL·E 2 struggles to accurately represent implied actions; e.g., in this particular case DALL·E 2 often includes the apple but frequently fails to capture the eating (Fig. 5, bottom). In addition, in half of the outputs, the semantic relations are reversed (i.e., the woman has a sandwich). When the grammar becomes complicated, the generated images often appear to be almost random combinations of the words that make up the prompt.

Mixing both negation and ellipsis causes further problems: 'The man eats pizza but the woman does not [~~eat pizza~~]' results in inaccurate depictions with the exception of one (Fig. 6). Meanwhile, structures involving a degree of semantic 'coercion' that also include ellipsis, as when we manipulate a verb such as 'start' to mean both make/eat and read/write, as in 'The girl starts a sandwich and the boy [~~starts~~] a book', also result in inaccurate depictions (Fig. 6).  In contrast, 3- to 5-year-olds reliably comprehend these various types of ellipsis at ceiling level performance.[22]

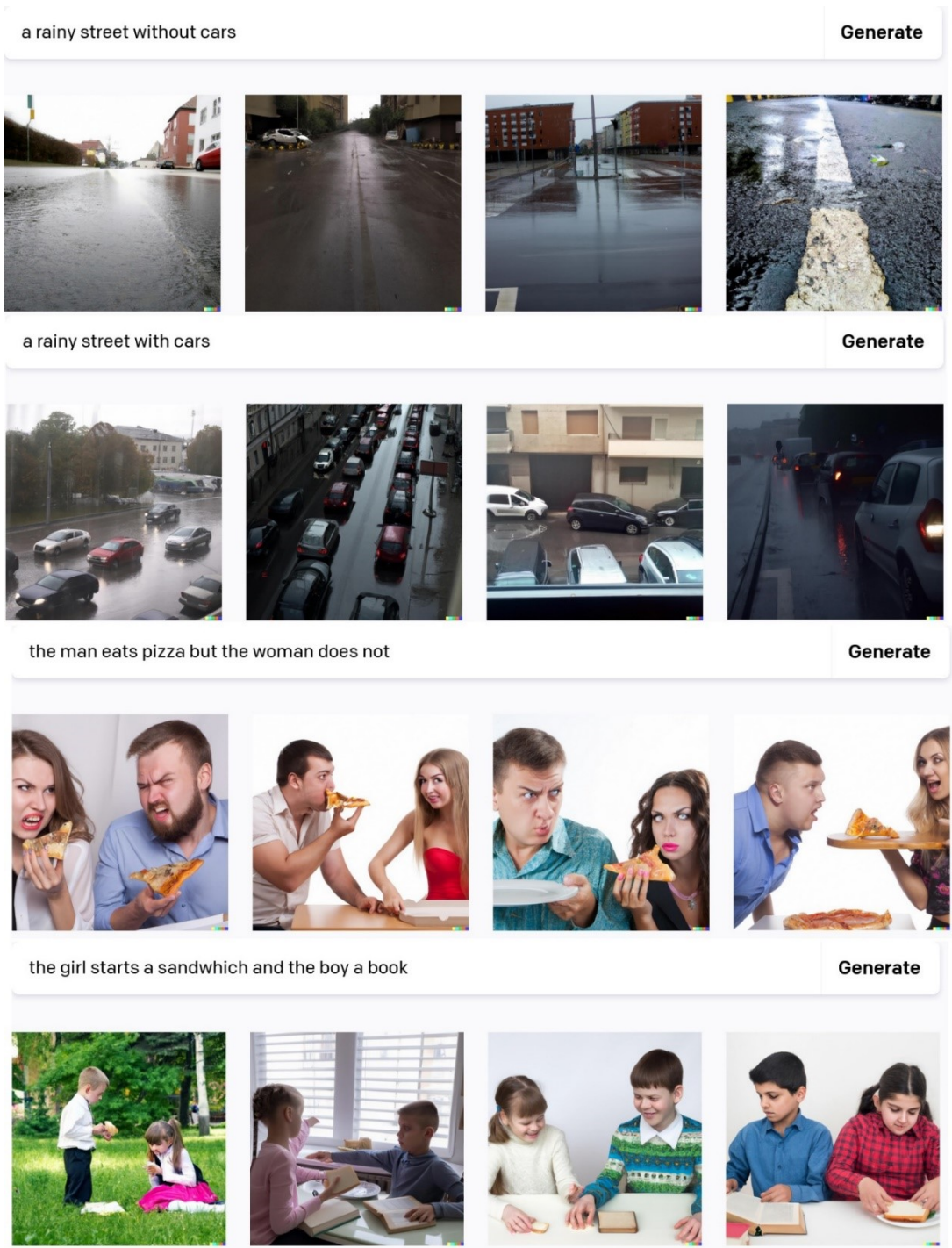

**Figure 6**: The performance of DALL·E 2 for a relational meaning both without negation (top) and with negation (second), and mixed negation-ellipsis (third) and coercion-ellipsis (last).

*Ambiguity*

Here, we examined prepositional phrase ambiguities ('The man saw the boy in his car', in which the man could be in the car, or only the boy, or both the man and the boy) and reduced relative clause ambiguities ('The boy saw the girl using a magnifying glass', in which either the boy or the girl could be using the instrument). We also examined ambiguities in which one of the parses violated common sense ('The man saw the lion with binoculars') (Fig. 7).

As with other cases we examined, performance was poor. For the prepositional phrase ambiguity, one representation correctly depicts one interpretation (man outside, boy inside), while another correctly depicts another (man and boy both inside), however half of the representations failed to depict a man, a boy, and an action that can be saliently construed as the one seeing the other.

For the common sense violation, only one representation appears to have depicted a man using binoculars to inspect a lion, and even here the binoculars are not clearly directed at the lion. For the reduced relative clause structures, three of the images depict an accurate image of a boy looking at a girl who is using the magnifying glass, but we do not get an image of the alternative parsing.

Overall, the majority of images failed to depict basic grammatical intuitions. 7-year-olds are known to use structural information to resolve syntactic ambiguity,[23] and more recent work has shown that even 3- to 5-year-olds can readily generate both parses for prepositional and reduced relative clause ambiguities.[24]

The lack of compositionality is also clear in the case of indefinite pronouns. While cases like 'There are three boys and each is wearing a hat' are represented well in only half of the representations (with the other images only showing two hats), cases that involve inter-object distinctions break down: 'Two cars each painted a different color' uniformly generates representations in which individual cars exhibit the property of being 'painted a different color', with the scope of the pronoun not being inferred. Only one representation accurately depicts two cars painted different colors (fuchsia and yellow). Omitting the pronoun, the string 'Two cars painted a different color' produces similar results, suggesting structural and semantic aspects of pronominal meaning are not generated. Note that this string does not say 'painted different colors', which also permits a mixed intra-object color scheme, it rather explicitly asks for discrete color schemes applied to both objects ('*a* different color', not '*many* different colors'), and yet we mostly get multicolored cars. Despite common *the*-overuse, 3- to 5-year-olds can accurately use indefinite pronouns.[25]

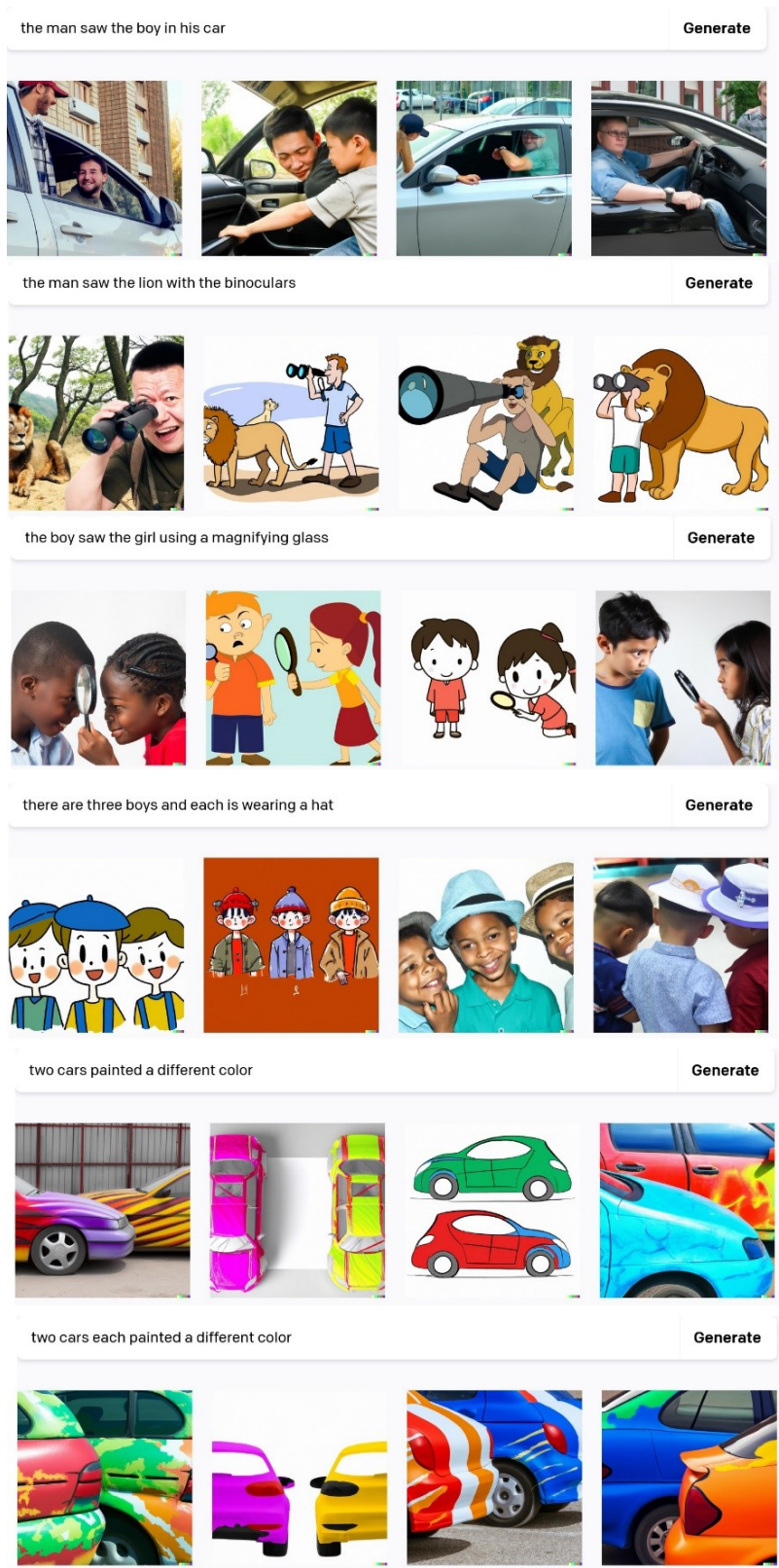

**Figure 7**: The performance of DALL·E 2 for structural ambiguity and indefinite pronouns.

***Summary***

On every facet of grammar that we studied, we found poor performance (Fig. 8); the full set of prompts and results are given in the Appendix.

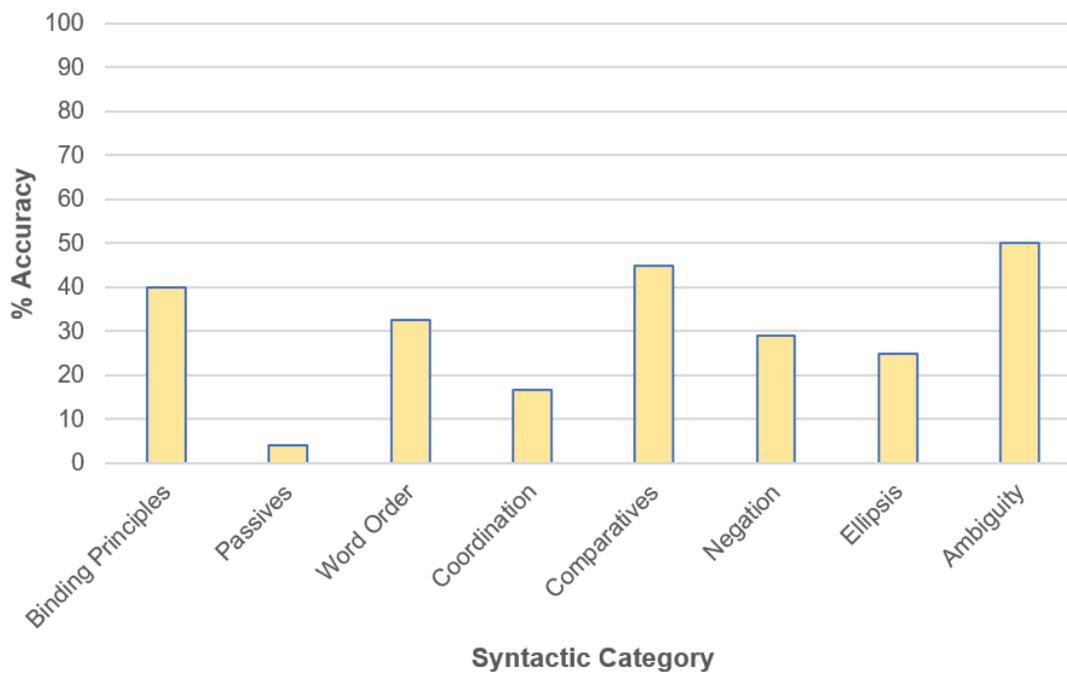

**Figure 8**: Accuracy across all tested syntactic categories.

## Discussion

Much recent research has aimed to show that some aspects of human language acquisition can be reduced to representations that are present in other species or, more recently, in AI systems. Campbell's monkeys possess a syntactic rule for combining different calls to produce sequences of more complex calls.[26-27] AI programs can generate results that show some understanding of grammatical rules into correctly translating some aspects of 'an elephant is behind a tree' (vs. 'a tree is behind an elephant').[6] However, more complex structure poses challenges.[28] Passives, double object constructions, and coordinated structures are early acquired by humans, but their parsing is problematic for systems like DALL·E 2.

Here, in the first systematic test of DALL·E 2 across a wide range of linguistic phenomena, we find consistent evidence that compositionality, one of the core properties of language discussed from Frege and Descartes to contemporary linguistics, is lacking.

Despite recent celebration of its linguistic abilities, DALL·E 2 generally performs at or near chance on syntax to semantics mappings that are routinely mastered by young children. The prompt-to-image translation of longer sentences often produces an outcome in which only some lexical items are drawn for interpretation, without a faithful compositional representation of their parts.

What is needed is not just linguistic input, but some *mapping* of linguistic structures onto some independent conceptual system, which has its own sets of compositional and inference rules, along with notions like truth/false, and faction/fiction, and cognitive models of the world[29-30]; synthesizing text alone is not necessarily in direct service to building a realistic model of human syntax and semantics.[31] In our view, all the recent attention that has been placed on predicting sequences of words has come at the expense of developing a theory of how such processes might culminate in cognitive models of the world, and how syntax serves to regulate form-meaning mappings.[32-33] For example, a recent account[34] claims that language models represent 'conceptual role' meanings because these can be inferred from relationships between internal representational states. Our results shows that such representations, to the extent that they exist at all, do not suffice. Humans can recruit language to refer to distinct mental faculties simultaneously (a 'newspaper' can be a physical object, or an abstract institution), and the regulation of semantics by these types of linguistic inputs remains a topic of ongoing research.[35-36] The very notion of 'conceptual role' presupposes a mapping to conceptual structures, and cannot be inferred purely from co-occurrence statistics of words within texts; current approaches do not seem to address this (i.e., what are the conceptual roles, and how do we know that common roles will be converged on and not some possible and computable role that does not exist for human language?).

Future work could move beyond simple pass/fail assessments, yet at this stage we aimed to explore the potential deficiencies of DALL·E 2 as a matter of error type, rather than to specify the degree of error across different syntactic phenomena. We also note, as have others who have explored this topic empirically,[9] that there are clear methodological issues concerning what level of participant agreement constitutes a 'success' for DALL·E 2 (40%? 70%?), instead of being more of a reflection of any number of independent participant preferences and biases, or participants being too generous with visually ambiguous images. Future work might examine other measures, e.g., by having participants freely describe images, or respond to a forced-choice decision between numerous linguistic descriptions of images.

Structure-dependence of rules are known by 18-month-old infants, and DALL·E 2 seems unable to implement this basic, structural property underlying most rules of syntax.[37] More general semantic deficits become salient here, increasingly discussed in the literature, pertaining to the (in)ability of transformer models and large language models to evaluate lexical inputs onto some general cognitive model of the world. To have a notion of reality, one might argue that we also need to have a notion of truth/false, and faction/fiction, from which to compare.[38] In addition, the subtle relationship between grammar and more general semantic and rules and comprehension strategies calling upon pragmatics and world knowledge needs to be addressed.[39-40]

It is currently vogue to attribute aspects of intelligence to deep learning models and large language models, and to dismiss traditional concepts from linguistics such as competence, innateness, non-linguistic, and unboundedness as outdated. In our view, this is a mistake. Attending to fundamental distinctions from linguistic theory, such as the distinction between computational competence and performance, structure-dependence of rules, or the explanatory power of compositionality, can contribute positively to building more robust, useful and trustworthy AI models.[41] This can, in turn, serve to reduce the reproduction of harmful biases and toxic speech commonly found in various language models, without losing sight of the clear importance of statistical learning even in some core syntactic processes[42-43].

Given the staggering mass of data DALL·E 2 has been trained on, we suggest that the grammatical failures it exhibits are not merely due to details of technical implementation, but rather reflect a qualitative difference between the computational basis of organic human language processing and the computational principles underlying current AI systems.

## Methods

We identified 8 grammatical phenomena that are pervasive in human language. We then translated them into natural language prompts that were given to DALL·E 2. In the Results, we present the first set of prompt-to-image translations returned for each case (without any filtering or selection procedure). These tests were run in August and October 2022. We then replicated all results, and supply the full set of tested materials and the outputs are made available as a benchmark for future testing in the Appendix.

# Appendix

We provide the full set of prompts and results (Table 1). The test set consists of 40 sets of prompts (8 syntactic constructions with 5 examples each). The accuracy of the algorithmic performance is ranked per set/output string. Figure 1 (main text) presents the overall algorithmic accuracy per tested category.

**Table 1**: Prompts and outputs.

| Binding Principles | | | |
|---|---|---|---|
| Prompt | Output | Acc. | Justification |
| 1. The man paints a picture of him/himself. | 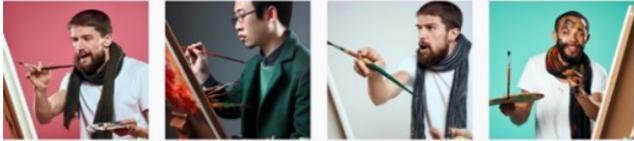 | 0% | The top 4 images depict an action that matches the prompt, but do not show the target of painting. The images in the second row either fail to capture the reflexive, or they interpret it as the man painting himself (vs. painting a picture of himself) |
| 2. The woman paints a portrait of her/herself. | 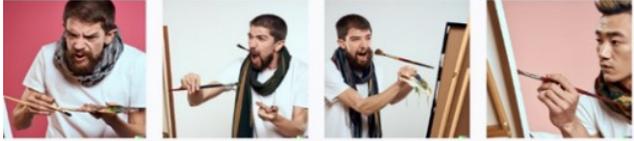 | 12.5% | Two of the images in the top row fail to show the agent. The images in the second row either fail to capture the reflexive, or they interpret it as the woman painting (towards) herself. |
| 3. The boy looks at a picture of him/himself. | 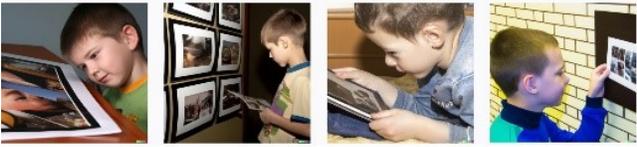 | 75% | One image shows the boy looking at the mirror (bottom #3), another does not clearly depict a male agent in the image (bottom #1). |
| 4. The young lady looks at a picture of her/herself. | 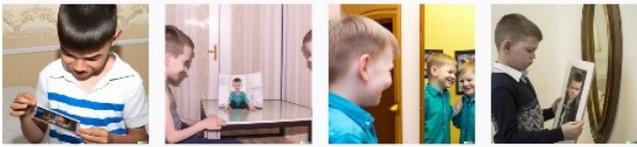 | 75% | One image shows the girl looking at the mirror, while another does not show the picture content. |
| 5. The man takes a picture of him/himself. | 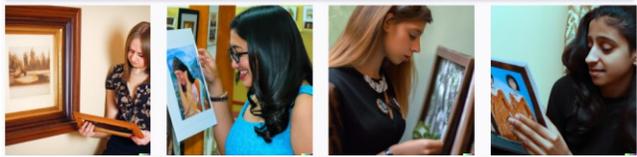 | 37.5% | In the first 3 images the target is not shown. In the 4th image, neither the target nor the action is shown. In the last image, the action is not shown. |
| **Passives** | | | |
| 6. The woman broke the vase/The vase was broken by the woman | 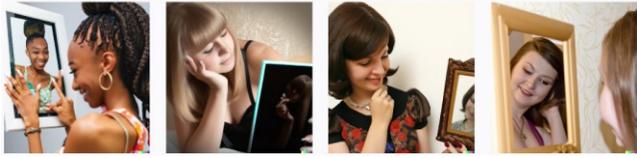 | 0% | The top row images do not show an act of breaking or the aftermath of a breaking event. The second-row images either do not show a woman or a broken vase. |
| 7. The plate was broken by the woman. | 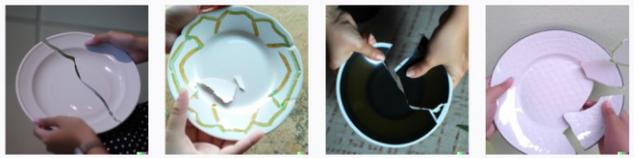 | 0% | No image shows a woman. |
| 8. The glass was broken by the man. | 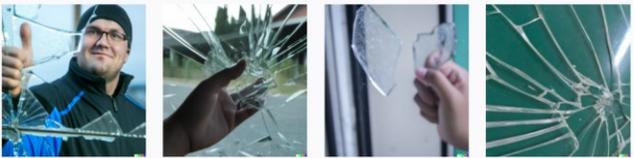 | 25% | One image shows a man. |
| 9. The jar was broken by the man. | 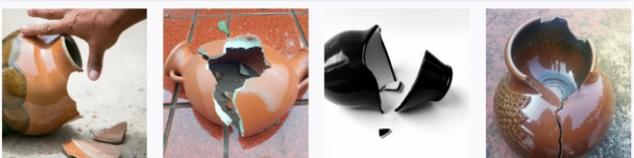 | 0% | No image shows a man. |
| 10. The flowerpot was broken by the man. | 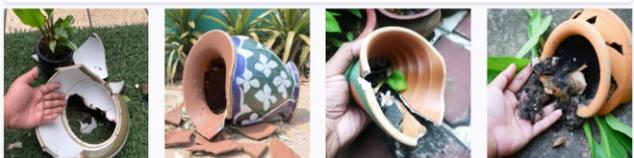 | 0% | No image shows a man. One image does not show a broken flowerpot. |

**Word order**

| | | | |
|---|---|---|---|
| 11. The dog is chasing the man/The man is chasing the dog | 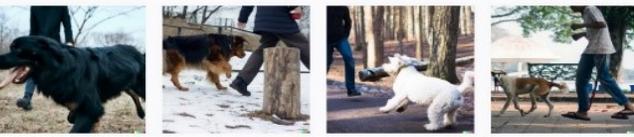 | 12.5% | In the second image, a dog is chasing a man. In the rest of the images, either there is no act of chasing, or the thematic roles are reversed. |
| 12. The man gave the letter to the woman/The man gave the woman the letter | 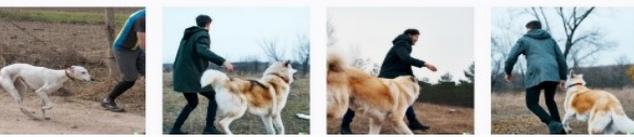 | 67.5% | Images 2 and 4 in the top row and 1, 2, 3 in the second row are accurate depictions of the prompt. |

| 13. The man is watering the plant/The plant is watering the man | 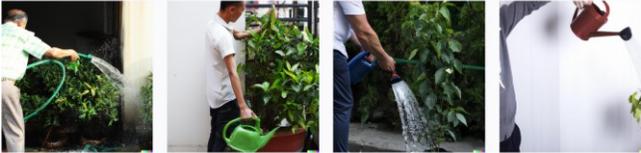 | 50% | All images in the top row are accurate depictions of the prompt. |
| 14. The mother combs the boy/The boy combs the mother | 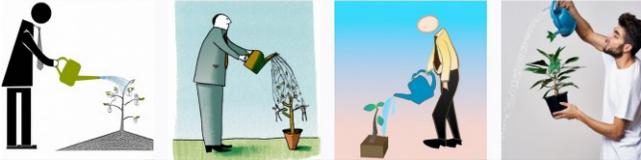 | 37.5% | Images 2, 3, and 4 in the top row show the target Agent, Patient, and action. |
| 15. The man gave the comb to the woman/The man gave the woman the comb | 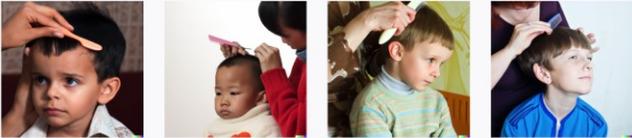 | 0% | The comb is in the hands of the man in all images. |
| **Coordination** | | | |
| 16. The man is drinking water and the woman is drinking orange juice | 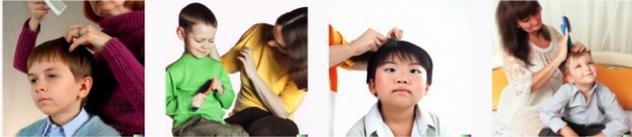 | 0% | In all images, the man is drinking orange juice. |
| 17. The woman is eating red apple and the man is eating a green apple | 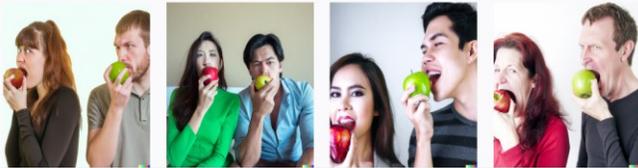 | 100% | |
| 18. The cat is wearing two red socks and the dog is wearing one red sock | 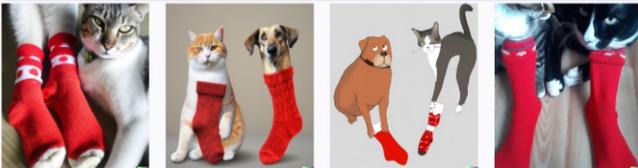 | 0% | No image shows a cat with two red socks. |
| 19. The boy wears a red hat and the girl wears a blue tie | 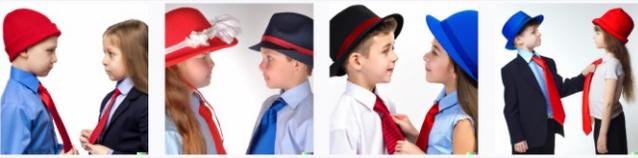 | 0% | No image shows a girl wearing a blue tie. |
| 20. The woman is washing the dishes and the man is washing the floor | 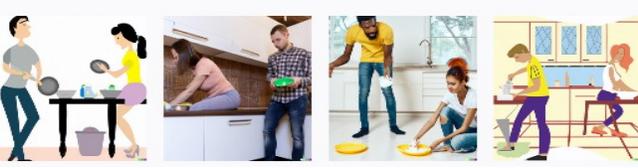 | 0% | No image shows a man washing the floor. |
| **Comparatives** | | | |
| 21. The bowl has more cucumbers than strawberries | 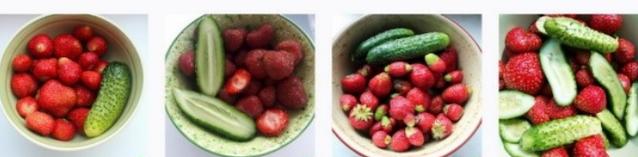 | 0% | All images show more strawberries than cucumbers. |
| 22. The bowl has fewer strawberries than cucumbers | 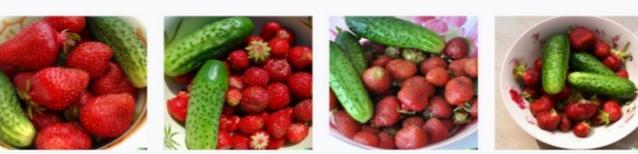 | 0% | All images show more strawberries than cucumbers. |
| 23. The plate has more peas than carrots | 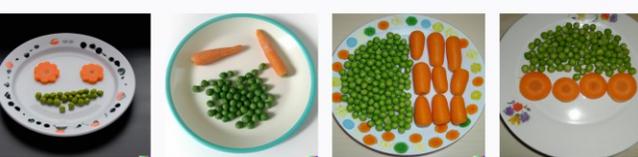 | 75% | Image 3 has a less than clear depiction of comparative size. |
| 24. The plate has fewer carrots than peas | 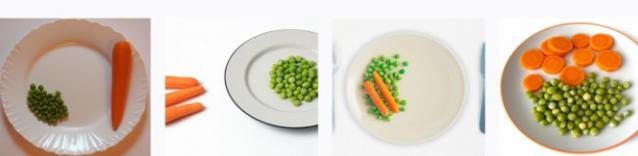 | 75% | |

| 25. The plate has more than seven eggs. | 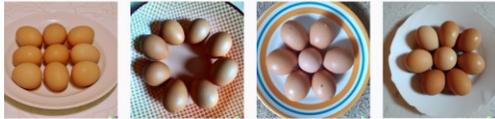 | 75% | Images 1, 2, and 4 show more than seven eggs. |
| --- | --- | --- | --- |
| **Negation** | | | |
| 26. A tall woman without a handbag. | 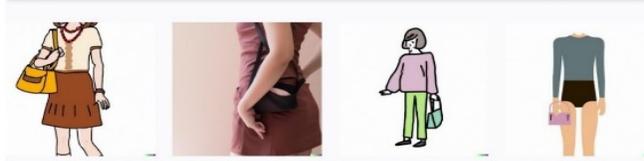 | 0% | All images show a handbag. |
| 27. A man with a red sweater and blue sweater and he is not wearing the former. | 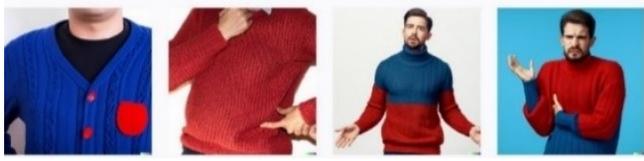 | 0% | No image shows two sweaters. |
| 28. A rainy street with/without cars. | 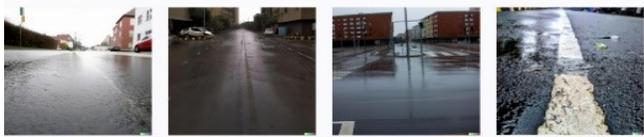 | 50% | Counting the first row as the target (i.e., negation), only images 3 and 4 do not show any cars. |
| 29. A boy with a green t-shirt without red buttons | 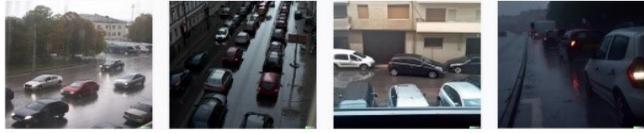 | 75% | Images 1, 2, and 4 are correct. |
| 30. A tall tree, not green or black. | 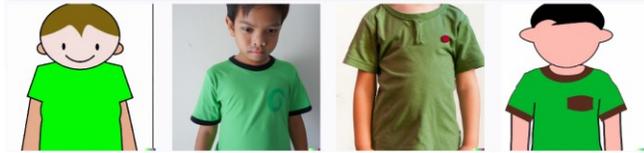 | 0% | All images show green trees. |
| **Ellipsis** | | | |
| 31. The man is eating a sandwich and the woman an apple. | 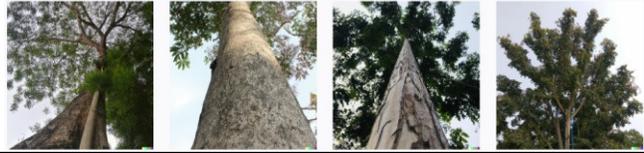 | 50% | Images 2 and 3 match the prompt. |
| 32. The man eats pizza but the woman does not | 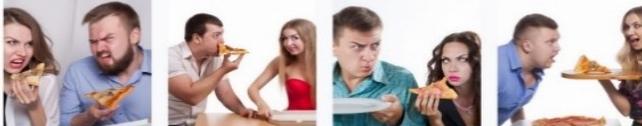 | 25% | Image 2 shows the action of eating with the correct Agent. |
| 33. The girl starts a sandwich and the boy a book. | 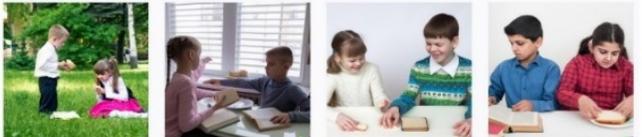 | 50% | Image 1 does not depict starting a book; 3 does not unambiguously depict the objects. |
| 34. The man drinks water and the woman orange juice. | 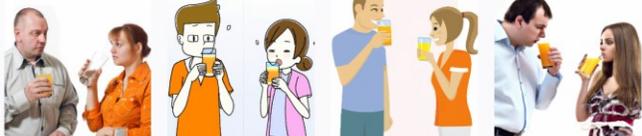 | 0% | No image shows a man drinking water. |
| 35. The woman wears a blue shirt, but the man does not. | 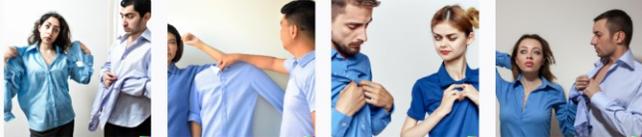 | 0% | All images show a man wearing a blue shirt. |
| **(Structural) Ambiguity** | | | |
| 36. The man saw the boy in his car. | 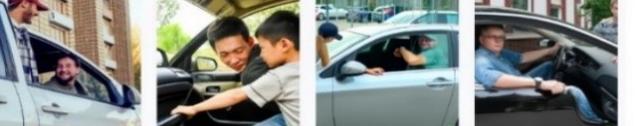 | 50% | Images 2 and 3 correctly depict one of the two meanings of the prompt. |
| 37. The man saw the lion with the binoculars. | 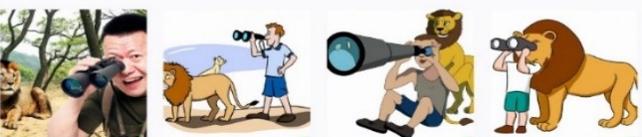 | 25% | Only image 2 shows a man using binoculars to inspect a lion. |
| 38. The boy saw the girl using a magnifying glass. | 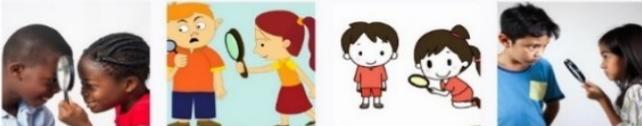 | 75% | Images 1, 3, and 4 show a boy looking at a girl who is using the magnifying glass. No image shows the alternative parsing. |
| 39. There are three boys and each is wearing a hat. | 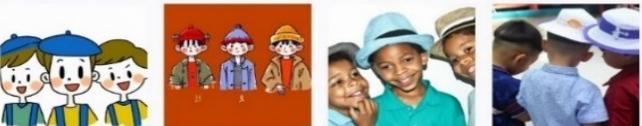 | 50% | Images 2 and 3 show all boys wearing a hat. |
| 40. Two cars painted different colors. (vs. (each) a different color) | 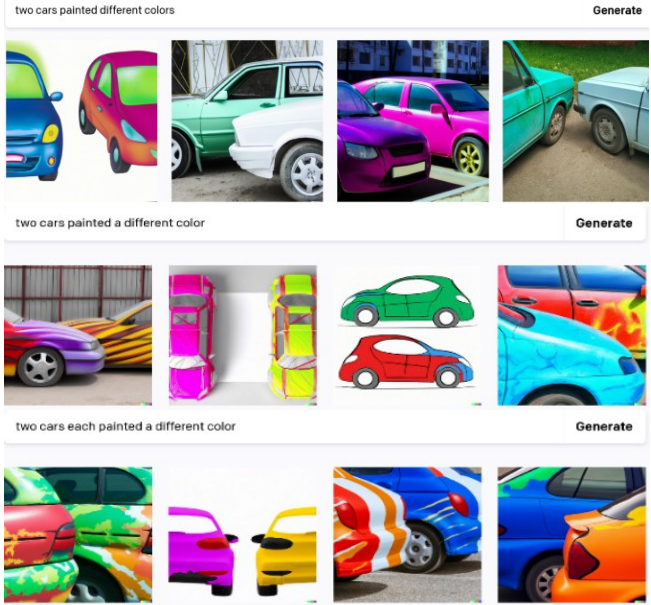 | 50% | The top row images are correct, and two of the bottom rows depict distinct colors. |